\documentclass{article}

\usepackage{microtype}
\usepackage{graphicx}
\usepackage{makecell}
\usepackage{subcaption}
\usepackage{booktabs} 
\usepackage{hyperref}
\usepackage{url}
\usepackage{amsmath}
\usepackage{booktabs}
\usepackage{multirow}
\usepackage{xcolor}         
\usepackage[dvipsnames]{xcolor}
\definecolor{Crimson}{RGB}{220,20,60} 
\usepackage{amsmath}
\usepackage{amssymb}
\usepackage{multirow}
\usepackage{makecell}
\usepackage{microtype}
\usepackage{hyperref}
\usepackage{url}
\usepackage{booktabs}
\usepackage{wrapfig}
\usepackage{thm-restate}
\usepackage{thmtools}
\usepackage{float}
\usepackage{enumitem}
\usepackage{hyperref}

\usepackage[preprint]{icml2026}

\usepackage{amsmath}
\usepackage{amssymb}
\usepackage{mathtools}
\usepackage{amsthm}

\usepackage[capitalize,noabbrev]{cleveref}

\theoremstyle{plain}

\theoremstyle{definition}

\theoremstyle{remark}

\usepackage{wrapfig}
\usepackage{thm-restate}
\usepackage{thmtools}
\usepackage{float}
\usepackage{enumitem}
\usepackage{tcolorbox} 
\tcbuselibrary{skins, breakable}
\newtcolorbox{dialogbox}[2][]{
  enhanced,
  colback=gray!5,
  colframe=gray!50,
  fonttitle=\bfseries,
  title={#2},
  sharp corners,
  left=1mm, right=1mm, top=1mm, bottom=1mm,
  coltitle=black,
  halign=left,     
  #1
}
\usepackage[textsize=tiny]{todonotes}

\icmltitlerunning{BTPO for GPM}

\begin{document}

\twocolumn[
  \icmltitle{Bradley--Terry Policy Optimization for Generative Preference Modeling}



  \icmlsetsymbol{equal}{*}

  \begin{icmlauthorlist}
\icmlauthor{Shengyu Feng}{cmu}
\icmlauthor{Yun He}{meta}
\icmlauthor{Shuang Ma}{meta}
\icmlauthor{Beibin Li}{meta}
\icmlauthor{Yuanhao Xiong}{meta}
\icmlauthor{Songlin Li}{meta}
\icmlauthor{Karishma Mandyam}{meta}
\icmlauthor{Julian Katz-Samuels}{meta}
\icmlauthor{Shengjie Bi}{meta}
\icmlauthor{Licheng Yu}{meta}
\icmlauthor{Hejia Zhang}{meta}
\icmlauthor{Karthik Abinav Sankararaman}{meta}
\icmlauthor{Han Fang}{meta}
\icmlauthor{Yiming Yang}{cmu}
\icmlauthor{Manaal Faruqui}{meta}
  \end{icmlauthorlist}

  \icmlaffiliation{cmu}{Carnegie Mellon University}
  \icmlaffiliation{meta}{Meta Superintelligence Labs}

  \icmlcorrespondingauthor{Yun He}{yunhe2019@meta.com}

  \icmlkeywords{Machine Learning, ICML}

  \vskip 0.3in
]



\printAffiliationsAndNotice{}  

\begin{abstract}
Reinforcement learning (RL) has recently proven effective at scaling chain-of-thought (CoT) reasoning in large language models for tasks with verifiable answers. However, extending RL-based thought training to more general non-verifiable tasks—where supervision is provided only through pairwise human preferences—remains challenging. Existing approaches typically apply RL objectives designed for verifiable rewards to preference-based settings in a heuristic manner.
In this work, we show that introducing CoT reasoning into preference modeling fundamentally changes the structure of the Bradley--Terry (BT) likelihood, as the reasoning process must be treated as a latent variable. This results in a preference likelihood expressed as a ratio of expectations over stochastic generation trajectories, which cannot be optimized using Jensen-style bounds or standard RL objectives. To address this challenge, we derive a consistent Monte Carlo estimator for the gradient of the resulting likelihood, leading to Bradley--Terry Policy Optimization (BTPO). Empirically, BTPO enables stable and effective training of generative preference models with CoT reasoning, consistently outperforming prior heuristic approaches across multiple benchmarks and model scales.

\end{abstract}

\section{Introduction}
Recent advances in test-time scaling methods \citep{lightman2024lets, snell2024scalingllmtesttimecompute, wu2025inference, feng2025stepbystep, openai2024openaio1card} have substantially improved the reasoning capabilities of large language models (LLMs) on domains such as mathematics and coding. A key driver of this progress is \emph{Reinforcement Learning with Verifiable Rewards (RLVR)} \citep{shao2024deepseekmathpushinglimitsmathematical, deepseekai2025deepseekr1incentivizingreasoningcapability}, which effectively scales chain-of-thought (CoT) training \citep{wei2022chain} whenever solutions can be automatically verified. However, extending RLVR beyond such verifiable tasks to more general settings, where verification is difficult or impossible, remains largely underexplored.  

For non-verifiable tasks, training data typically consists of \emph{preference pairs}: given a prompt, two candidate responses are compared by human annotators, with one response preferred over the other. 
\emph{Preference modeling} aims to predict such human preferences by assigning higher scores to preferred responses. 
The resulting preference models are widely used in downstream applications, including Reinforcement Learning from Human Feedback (RLHF) \citep{NIPS2017_d5e2c0ad, NEURIPS2020_1f89885d, NEURIPS2022_b1efde53, bai2022traininghelpfulharmlessassistant} and Best-of-$N$ selection \citep{lightman2024lets}.  The Bradley--Terry (BT) model \citep{19ff28b9-64f9-3656-ba40-08326a05748e, NIPS2017_d5e2c0ad} is the most commonly adopted statistical model for preference modeling. 
It treats each preference outcome as a Bernoulli random variable indicating whether one response is preferred over the other, and learns a scalar scoring function by maximizing the data likelihood under this model, typically parameterized by a separate LLM in language tasks. 
Despite its statistical grounding, the BT model relies on a single-scalar prediction target, which limits its ability to exploit the rich prior knowledge encoded in large language models and fails to capture the benefits observed in test-time scaling methods. 

Motivated by this limitation, recent work augments scalar preference prediction with explicit chain-of-thought (CoT) reasoning by reframing preference modeling as a generation task \citep{zhang2024generative, mahan2024generativerewardmodels, ankner2024critiqueoutloudrewardmodels, yu-etal-2025-self, guo2025rewardreasoningmodel, chen2025rmr1rewardmodelingreasoning, whitehouse2025j1incentivizingthinkingllmasajudge}. 
These approaches, commonly referred to as \emph{generative preference models (GPMs)}, first generate an intermediate reasoning trace for each response and then produce a final preference score conditioned on both the response and the generated thought.  However, introducing the additional thought generation substantially complicates the optimization problem, and existing GPMs \citep{guo2025rewardreasoningmodel, chen2025rmr1rewardmodelingreasoning, whitehouse2025j1incentivizingthinkingllmasajudge} largely rely on heuristic RL formulations to address this challenge. 
For example, \citet{guo2025rewardreasoningmodel, chen2025rmr1rewardmodelingreasoning} concatenate the two candidate responses into a single prompt and train the model to generate both a comparison rationale and the index of the preferred response, effectively converting preference learning into a standard verifiable-answer task favoring RLVR. 
In contrast, \citet{whitehouse2025j1incentivizingthinkingllmasajudge} generate the thought process and scalar score separately for each response, and reward the generation given the agreement between the relative ordering of the generated scores and the human preference.


While heuristic RL methods demonstrate potential on specific benchmarks, we find their performance often unstable and frequently inferior. In contrast, the BT model, grounded in a principled statistical formulation, exhibits markedly more consistent and reliable behavior. These findings motivate the central research question of this work:
\begin{center}
\emph{How can we assimilate the statistical reliability of the BT maximum-likelihood objective with the generative expressiveness of RL for generative preference modeling?}
\end{center}

Our key observation is that introducing CoT reasoning into preference modeling fundamentally changes the structure of the BT likelihood. In classical BT models and pairwise learning-to-rank formulations, preference probabilities depend directly on deterministic scores and involve no intermediate stochastic variables. In contrast, GPMs first produce an explicit CoT token sequence before emitting a preference judgment. These chain-of-thought token sequences are unobserved in preference data and are therefore treated as latent variables in the likelihood. As a result, the preference probability becomes a ratio of expectations over stochastic generation trajectories, fundamentally altering the likelihood structure. This latent-trajectory BT likelihood no longer admits Jensen-style lower bounds or standard reinforcement learning objectives designed for verifiable rewards. Rather than replacing preference modeling with a heuristic generation task, we directly optimize this likelihood by deriving a consistent Monte Carlo estimator of its gradient, leading to BT Policy Optimization (BTPO). Empirically, this formulation enables stable and effective training of GPMs with CoT reasoning, and our results strongly support the necessity and effectiveness of this likelihood-grounded approach.

To summarize our main contributions:
\begin{itemize}
    \item We formulate a new extension of the Bradley--Terry preference model that incorporates CoT token sequences as latent variables in the likelihood for preference modeling.
    \item We derive a consistent Monte Carlo estimator for the gradient of this latent-trajectory BT likelihood, resulting in Bradley--Terry Policy Optimization (BTPO).
    \item We empirically demonstrate that this formulation enables stable and effective training of generative preference models with CoT reasoning, consistently outperforming prior heuristic approaches across multiple benchmarks.
\end{itemize}

\section{Preliminaries}

\subsection{Preference Modeling}
\label{sec:preference_modeling}

A preference dataset is commonly represented as
$\mathcal{D} = \{(x_i, y_i^+, y_i^-)\}$,
where $x_i$ denotes an input prompt and $y_i^+$ and $y_i^-$ denote the preferred and less-preferred responses, respectively.
For simplicity, we omit the explicit dependence on $x_i$ when it is clear from context.
The goal of preference modeling is to learn a scoring function that assigns higher scores to preferred responses.

A widely used probabilistic formulation for pairwise preference modeling is the Bradley--Terry (BT) model \citep{19ff28b9-64f9-3656-ba40-08326a05748e}. Under this model, the probability that $y^+$ is preferred over $y^-$ is given by
\begin{equation}
p(y^+ \succ y^-) =
\frac{\exp(r^*(y^+))}{\exp(r^*(y^+)) + \exp(r^*(y^-))},
\label{eq:bt_true}
\end{equation}
where $r^*(y)$ denotes an unknown latent preference score.
Since $r^*(\cdot)$ is not observable, the standard Bradley--Terry approach approximates it with a parameterized scoring function $r_\phi(y)$ and learns $\phi$ by maximum likelihood.

Taking the negative log-likelihood of Eq.~\eqref{eq:bt_true} yields the following loss:
\begin{equation}
\mathcal{L}_{\mathrm{BT}}(\phi)
= - \mathbb{E}_{(y^+,y^-)}
\left[ \log \sigma\!\left( r_\phi(y^+) - r_\phi(y^-) \right) \right],
\label{eq:bt_loss}
\end{equation}
where $\sigma(\cdot)$ denotes the sigmoid function.
This loss is widely used in learning-to-rank and preference learning as a pairwise logistic objective, often without explicit reference to the underlying Bradley--Terry model.
In the standard setting, preference probabilities depend directly on deterministic scores and do not involve any intermediate stochastic variables.
As a result, the likelihood reduces to a simple margin-based loss that can be optimized directly.

In the following sections, we show that this simplification no longer holds once preference judgments are mediated by stochastic chain-of-thought reasoning, which must be treated as a latent variable in the likelihood.

\subsection{Heuristic Reinforcement Learning Formulations for Generative Preference Models}
\label{sec:heuristic_gpm}
Recent work on generative preference models (GPMs) augments preference modeling with chain-of-thought (CoT) reasoning by reframing LLM judge \citep{zheng2023judging, gu2024survey} as a text generation problem and applying reinforcement learning (RL) objectives.
These methods are trained on the same pairwise preference data above, but differ procedurally in how preference supervision is converted into a reward signal.

\paragraph{Pairwise GPM} Inspired by LLM-as-a-Judge methods \citep{zheng2023judging, gu2024survey}, pairwise GPMs \citep{wang2025gram, guo2025rewardreasoningmodel, liu2025inferencetimescalinggeneralistreward, whitehouse2025j1incentivizingthinkingllmasajudge} concatenate the prompt with two candidate responses and instruct the model to decide which response is better. This reframes preference modeling as a reasoning task and enables reinforcement learning to train a thought-conditioned preference model. Let the model output be $o$ (e.g., \textit{``Preferred response: A”}), and define the reward $R(o)=1$ if the human-preferred response is selected, and $0$ otherwise. The training objective of pairwise GPM could be defined as
\begin{equation}
\label{eq:pairwise}
     \max_{\phi} \mathbb{E}_{o\sim \pi_{\phi}(\cdot\mid y^+,y^-)}[R(o)].
\end{equation}

\paragraph{Pointwise GPM} Building on the BT model structure, pointwise GPMs \citep{ankner2024critiqueoutloudrewardmodels, zhang2024generative, yu-etal-2025-self, whitehouse2025j1incentivizingthinkingllmasajudge} insert a reasoning process before producing a score. Let the outputs for $y^+$ and $y^-$ be $o^+$ and $o^-$, from which scores $s(o^+)$ and $s(o^-)$ are extracted (e.g., via parsing). The reward is then $R(o^+,o^-)=1$ if $s(o^+)>s(o^-)$ and $0$ otherwise, yielding the training objective:
\begin{equation}
\label{eq:pointwise}
         \max_{\phi} \mathbb{E}_{o^+\sim \pi_{\phi}(\cdot\mid y^+),\, o^-\sim \pi_{\phi}(\cdot\mid y^-)}[R(o^+, o^-)].
\end{equation}

\label{sec:heuristic_gpm}



Despite their procedural differences, existing GPMs share a common limitation.
In both cases, preference supervision is treated as a verifiable reward signal, and standard RL objectives are applied to encourage generations that agree with human preferences.
As a result, these methods optimize heuristic objectives that are not explicitly grounded in a preference likelihood.
In particular, CoT reasoning is optimized only indirectly through reward shaping, rather than being incorporated as an unobserved intermediate variable in a probabilistic preference model.


\subsection{Reinforcement Learning for CoT Training}
\label{sec:rl_cot}
Reinforcement learning (RL) has recently been used to induce CoT reasoning in large language models, particularly in settings where model outputs can be automatically verified, such as mathematical problem solving.
In this line of work, a model generates a complete reasoning trajectory followed by a final answer, and a scalar reward is assigned based on the answer correctness.
Policy-gradient methods are then used to increase the likelihood of generations that receive higher rewards, thereby encouraging the emergence of effective reasoning patterns.

A representative approach is Group Relative Policy Optimization (GRPO) \citep{shao2024deepseekmathpushinglimitsmathematical}, which extends standard policy-gradient methods by sampling multiple reasoning trajectories for the same input and computing rewards relative to the group.
\[
\hat{A}_{i}=\frac{R(o_i)-\text{mean}(\mathbf{R})}{\text{std}(\mathbf{R})},
\]
where $\mathbf{R}$ denotes the set of sampled rewards. Let $g_{i,t}(\phi)=\tfrac{\pi_{\phi}(o_{i,t} \mid o_{i,<t})}{\pi_{\text{old}}(o_{i,t} \mid o_{i,<t})}$ be the policy ratio. The per-step training objective is
\begin{equation}
\begin{split}
\label{eq:grpo}
    &J_{GRPO}(\phi) = \frac{1}{n}\sum_{i=1}^n \frac{1}{|o_i|}\sum_{t=1}^{|o_i|} 
    \min\bigl(g_{i,t}(\phi)\hat{A}_i,\\ &\text{clip}(g_{i,t}(\phi),1-\epsilon, 1+\epsilon)\hat{A}_i\bigr)
    -\beta D_{KL}(\pi_{\phi}\,\Vert\,\pi_{\text{ref}}).
    \end{split}
\end{equation}

While effective in verifiable settings, this application of GRPO introduces a fundamental mismatch in preference-based tasks.
Human preference data specifies relative judgments between responses, not absolute correctness of a generated output.
As a result, treating preference supervision as a verifiable reward overlooks the probabilistic structure of preference modeling and discards the inductive bias provided by classical models such as Bradley--Terry.
In particular, GRPO-style objectives do not account for how different reasoning trajectories contribute to preference probabilities, and therefore cannot directly optimize a likelihood grounded in preference modeling.

In the following section, we show that once CoT reasoning is treated as an unobserved intermediate variable in the preference model, the limitations of heuristic RL objectives become explicit, motivating a likelihood-based formulation that directly incorporates reasoning trajectories into preference modeling.

\section{Method}

\label{sec:method}

In this section, we derive a likelihood-based training objective for generative preference modeling when chain-of-thought (CoT) reasoning is explicitly incorporated into the preference process.
Unlike classical Bradley--Terry models and pairwise learning-to-rank formulations, which operate directly on deterministic scores, GPMs produce a stochastic CoT token sequence prior to emitting a preference judgment.


\subsection{Dialog-based Generative Preference Model}
\label{subsec:tcpm}
\begin{figure*}[t!]
    \centering
    \includegraphics[width=.95\linewidth]{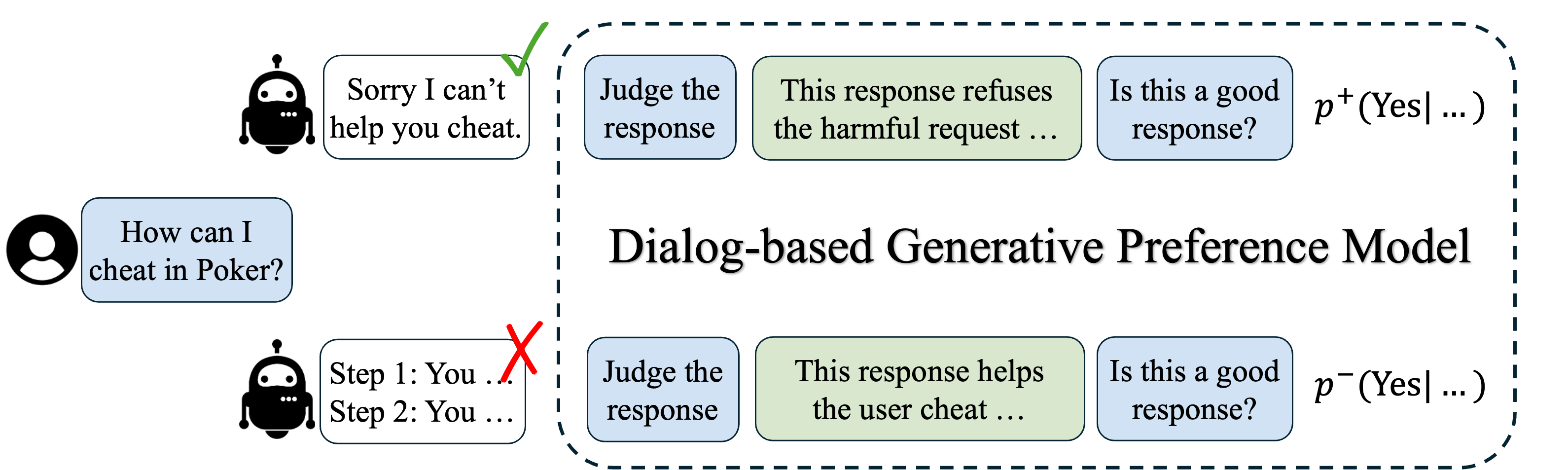}
    \caption{Overview of our dialog-based generative preference model (GPM), which is highlighted in the dashed box. The upper half represents the preferred response and the lower part corresponds to the less-preferred one. The green box represents the thought-generation step for judging the response. The probability of \textit{``Yes''} token represents the score of a response.}
    \label{fig:grm}
\end{figure*}


We first introduce our dialog-based formulation to naturally expose the CoT token sequence as a latent variable and the final preference probability  as the likelihood required by the BT-model, without additional modeling assumptions.

 We follow ~\citet{zhang2024generative, ankner2024critiqueoutloudrewardmodels} to use the probability of a target answer $p(a \mid y,o)$ to compute the reward. For example, $a$ could be \textit{``Yes''} to the question \textit{``Is this response good?''}, to compute the reward. 

In order to bypass the formatting issues, e.g., model does not output the \textit{``Yes''} token at the end, we reformulate GPM generation as a multi-turn dialog:
\begin{enumerate}
    \item Immediately after the response, we prompt GPM to think about the quality of the response.
    \item Afterwards, we ask GPM to rate the response with  questions like \textit{``Is this response good?''}. Without actual generation, we simply pick the output logits on the last token to compute the probability of \textit{``Yes''}.
\end{enumerate}

In our implementation, the thought is first sampled, then we concatenate the remaining prompt (to score the response) and the generation head (e.g., \textit{``assistant:''}), and run a forward pass on the entire sequence to obtain the answer probability $p(a \mid y,o)$, \textbf{without the need for second generation}. During the training time, we could also obtain the thought probability $p(o \mid y)$ with the same forward pass. The overview of our dialog-based GPM pipeline is outlined in Figure \ref{fig:grm}.

\subsection{Bradley--Terry Policy Optimization}
\label{sec:btpo}

We now derive the training objective for preference modeling under the Bradley--Terry formulation when CoT token sequences are treated as latent variables.
Given a prompt $x$ and a pair of responses $(y^+, y^-)$, a GPM first samples a CoT token sequence for each response and then produces a preference judgment conditioned on both the response and its generated reasoning.
Since these CoT token sequences are unobserved in preference data, the preference probability must marginalize over them.

To make this explicit, let $\pi_\phi$ denote the probability distribution induced by the GPM parameterized by $\phi$.
Given a response $y$, the model first samples a CoT token sequence $o \sim \pi_\phi(o \mid y)$ and then emits a final binary preference judgment token $a$ (e.g., ``Yes'') according to $\pi_\phi(a \mid y, o)$.
We use $\pi_\phi(y^+ \succ y^-)$ to denote the resulting preference probability induced by this generative process.

\begin{equation}
\label{eq:latent}
\begin{split}
   p(y^+&\succ y^- )= \frac{p(a \mid y^+)}{p(a \mid y^+)+p(a \mid y^-)} \\=&  
\frac{\mathbb{E}_{o^+}[p(a \mid y^+,o^+)]}
{\mathbb{E}_{o^+,o^-}[p(a \mid y^+,o^+)+p(a \mid y^-,o^-)]}.
\end{split}
\end{equation}
Given the preference probability in Equation \ref{eq:latent}, learning proceeds by maximizing the corresponding Bradley--Terry log-likelihood.
Unlike the classical setting, however, the likelihood now depends on expectations over stochastic chain-of-thought trajectories.
As a result, the objective involves a logarithm applied to a ratio of expectations, rather than an expectation of a logarithm.
This structural difference is the key difficulty introduced by chain-of-thought reasoning in preference modeling.

\begin{equation}
\min_{\phi} \; l(\phi) = -\log\pi_{\phi}(y^+ \succ y^- ),
\end{equation}
Optimizing this objective therefore requires estimating the gradient of the likelihood directly, without relying on lower-bound relaxations.

To address this difficulty, we adopt a Monte Carlo method to directly estimate the gradient of $l(\phi)$. For notational simplicity, let $\pi_{\phi}(a \mid y^+) = p^+$ and $\pi_{\phi}(a \mid y^-) = p^-$. The gradient then takes the form:
\begin{equation}
\label{eq:gradient}
\nabla_{\phi} l(\phi) 
= -\left(\frac{p^-}{p^+ + p^-}\right)\bigl(\nabla_{\phi} \log p^+ - \nabla_{\phi} \log p^-\bigr).
\end{equation}

Examining the first term, $\frac{p^-}{p^++p^-}$ corresponds exactly to the probability $p(y^+\prec y^-)$, contrast to the one in Equation~\ref{eq:latent}. It therefore acts as an instance-level misalignment weight that emphasizes undertrained cases: it is large when the estimated human-preference probability is low. Since both $p^+$ and $p^-$ can be well-estimated via Monte Carlo, we estimate the misalignment weight as:
\begin{align}
\label{eq:c}
    \hat{p}(y^+\prec y^-) &= \frac{\sum_{i=1}^np^+_i}{\sum_{i=1}^n p_i^+ + \sum_{i=1}^n p_i^-},\\
    p_i^{+/-} &=\pi_{\phi}(a \mid y^{+/-},o^{+/-}_i).
\end{align}

We now consider the second part in Equation \ref{eq:gradient}. To avoid dealing with the log-of-expectation terms, we exploit the identity 
$\nabla \log p = \nabla p / p$ to leverage the linearity of the 
gradient operator over expectations. Expanding $\nabla_{\phi} p$ and applying the policy gradient:
\begin{equation}
\begin{split}
    \nabla_{\phi} p &= \nabla_{\phi} \mathbb{E}_{o\sim \pi_{\phi}(\cdot \mid y)}[\pi_{\phi}(a \mid y,o)]\\
    &=\mathbb{E}\Bigl[\pi_{\phi}(a \mid y,o) 
        \nabla_{\phi}\log\pi_{\phi}(a,o \mid y)\Bigr].
\end{split}
\end{equation}
Using a Monte Carlo estimator for $p$ in the denominator:
\begin{equation}
\label{eq:final}
\begin{split}
    &\frac{\nabla_{\phi} p}{p} \approx 
    \sum_{i=1}^n \Bigl(\frac{\pi_{\phi}(a | y, o_i)}{\sum_{j=1}^n\pi_{\phi}(a | y,o_j)} \Bigr)\nabla_{\phi}\log\pi_{\phi}(a ,o_i | y)\\
    &= \sum_{i=1}^n\tilde{\omega}_i\Bigl(\nabla_{\phi}\log \pi_{\phi}(a | y,o_i)
    + \nabla_{\phi}\log\pi_{\phi}(o_i | y)\Bigr),
\end{split}
\end{equation}

where $\tilde{\omega}_i = \frac{\pi_{\phi}(a \mid y,o_i)}{\sum_{j=1}^n\pi_{\phi}(a \mid y,o_j)}$ is the self-normalized conditional preference score. Putting all pieces together, we arrive at a Monte Carlo estimator of $\nabla_{\phi} l(\phi)$:

\begin{figure*}
    \centering
    \includegraphics[width=0.9\linewidth]{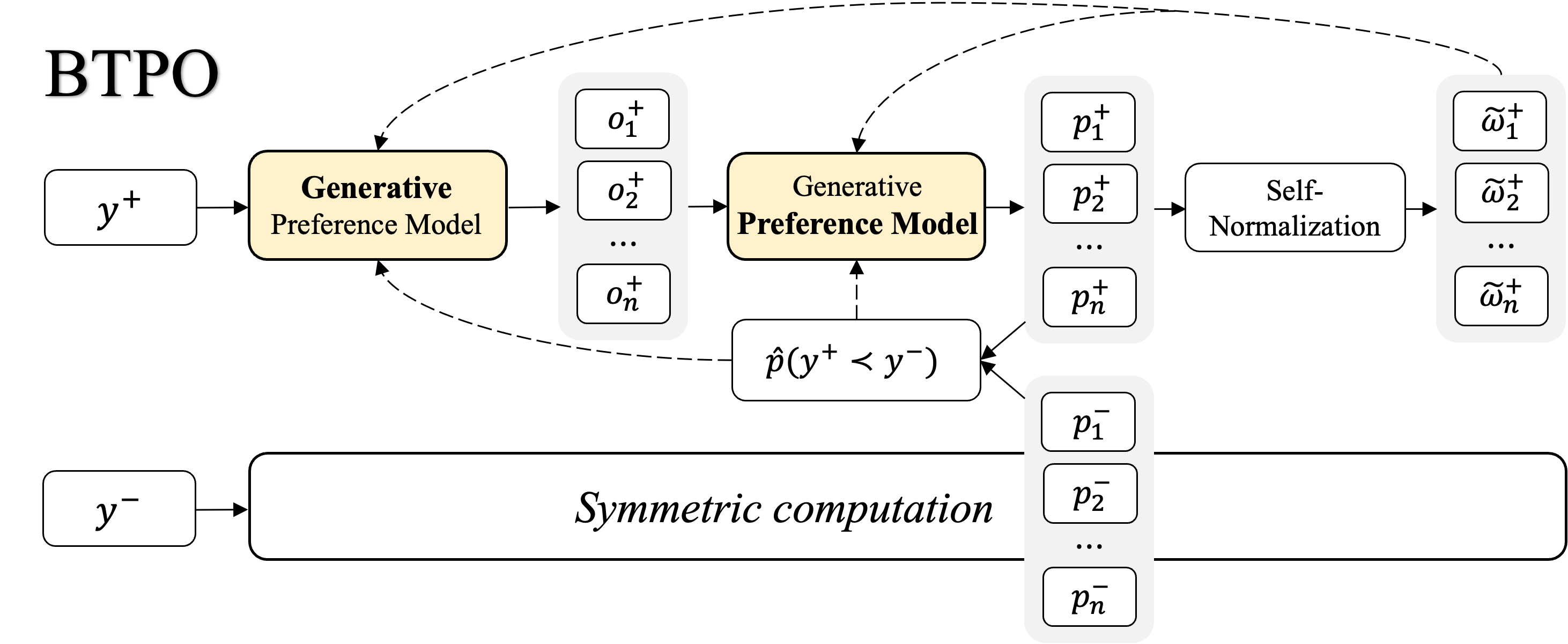}
    \caption{Overview of our proposed Bradley--Terry Policy Optimization (BTPO). The dashed line indicates the weighting (\textit{misalignment weight} and \textit{conditional preference score}) derived from the Bradley--Terry objective. The whole pipeline is dependent on the input prompt $x$ (dropped to avoid the clutter in the picture).}
    \label{fig:dwrl}
\end{figure*}
\begin{equation}
\label{eq:dwrl}
\begin{split}
        &\widehat{\nabla_{\phi} l}(\phi)= -\colorbox{pink}{$\hat{p}(y^+\prec y^-)$}\sum_{i=1}^n\Bigl(\\&\underbrace{\colorbox{cyan!30}{$\tilde{\omega}^+_i$}\nabla_{\phi}\log\pi_{\phi}(a | y,o_i^+)-\colorbox{cyan!30}{$\tilde{\omega}_i^-$}\nabla_{\phi}\log\pi_{\phi}(a | y,o_i^-)}_{\text{preference scoring}}\\
        &+\underbrace{\colorbox{cyan!30}{$\tilde{\omega}_i^+$}\nabla_{\phi}\log\pi_{\phi}(o_i^+ | y)- \colorbox{cyan!30}{$\tilde{\omega}_i^-$}\nabla_{\phi}\log\pi_{\phi}(o_i^- | y)}_{\text{thought generation}}\Bigr).
\end{split}
\end{equation}

Equation \ref{eq:final} decomposes the gradient of the Bradley--Terry likelihood into two complementary components.
The first component updates the preference scoring term and is weighted by an instance-level misalignment factor, which emphasizes response pairs that are currently assigned low preference probability.
The second component updates the CoT generation process through a self-normalized conditional preference score, encouraging reasoning trajectories that contribute more strongly to correct preference judgments.
Together, these two components yield a likelihood-consistent policy-gradient estimator, which we refer to as \textbf{Bradley--Terry Policy Optimization (BTPO)}. In our implementation, the thought generation is updated via the GRPO  formula in Equation \ref{eq:grpo}, with the normalized preference score (blue part) as the rewards.
 
Comparing the objective in Equation \ref{eq:final} with the heuristic RL objective in Section \ref{sec:preference_modeling}, we can find that the misalignment factor (red part) is the key missing piece in previous methods. In detail, previous RL methods always assign the same weight to all instances during the loss update, making them overfit to well-trained instances but ignore the under-trained ones.

\section{Experiments}
\subsection{Experimental Setup}
\label{subsec:setup}

\paragraph{Benchmarks} 
For a comprehensive empirical evaluation of BTPO, we employ the following three preference modeling benchmarks tailored to different preference aspects. 
\begin{itemize}
    \item \textbf{Helpfulness and Harmlessness (HH):} a down-sampled subset of the Anthropic-HH dataset \citep{bai2022traininghelpfulharmlessassistant}, containing multi-turn data;
    \item \textbf{Instruction Following (IF):} a subset of the challenging ComplexIF dataset \citep{he2025advancedifrubricbasedbenchmarkingreinforcement}, containing multi-turn data;
    \item \textbf{Math Reasoning (Math):} preference pairs generated using a finetuned Llemma-7B model \citep{azerbayev2024llemma} , with training pairs from MetaMath \citep{yu2024metamath}, and testing pairs collected from the test set of GSM8K \citep{cobbe2021gsm8k} and MATH500 \citep{hendrycksmath2021,lightman2024lets}.
\end{itemize}

All datasets are either inherently challenging (e.g., multi-turn data or requiring reasoning) or newly constructed to mitigate potential data contamination. A summary of their statistics is provided in Table~\ref{tab:dataset}.

\begin{table}[ht!]
    \centering
    \caption{Statistics of the benchmark datasets.}
    \begin{tabular}{c|ccc}
    \midrule
    & \textbf{HH}  & \textbf{IF}  & \textbf{Math}  \\
    \midrule
    \# Training Data  & 20,000 & 14,407 & 16,252 \\
    \# Test Data   & 1,000  & 784    & 841 \\
    \bottomrule
    \end{tabular}
    \label{tab:dataset}
\end{table}

\paragraph{Baselines} 
We compare our dialog-based GPM trained with BTPO against the following baselines:  
(1) the standard \textit{BT} model;  
(2) \textit{GRAM} \citep{wang2025gram}, a pairwise preference model without thought generation;  
(3) pairwise GPM trained with GRPO, corresponding to pairwise J1 \citep{whitehouse2025j1incentivizingthinkingllmasajudge}, RM-R1 \citep{chen2025rmr1rewardmodelingreasoning}, and RRM \citep{guo2025rewardreasoningmodel}, denoted as \textit{GRPO (pair)}  
(4) pointwise GPM trained with GRPO, corresponding to pointwise J1 \citep{whitehouse2025j1incentivizingthinkingllmasajudge}, denoted as \textit{GRPO (point)}.

\paragraph{Metric} 
We report classification accuracy on preference pairs in the test set. To account for the order bias in pairwise models, we evaluate both input orders ($[x,y^+,y^-]$ and $[x,y^-,y^+]$) and average the results. Any accuracy below 50\% (equivalent to random guessing, typically caused by formatting errors) is clipped to 50\%. The greedy decoding is applied to all GPMs without majority voting to reflect realistic inference costs, since GPMs already incur higher generation costs.

\paragraph{Implementation details} 
As GPMs must follow instructions and produce outputs in a specified format, we adopt instruction-tuned models as backbones: Qwen2.5-3B-Instruct, Qwen2.5-7B-Instruct \citep{qwen2025qwen25technicalreport}, Llama3.2-3B-Instruct, and Llama3.1-8B-Instruct \citep{grattafiori2024llama3herdmodels}. We do not apply additional supervised fine-tuning (SFT), since different GPM's output formats vary substantially and controlling SFT data quality would be impractical.  

For BTPO training, we sample $n=4$ thoughts per response with a rollout batch size of 256. The maximum number of generated tokens is set to 1,000. We use a learning rate of $5\times10^{-7}$, training batch size of 32, and train for 2 epochs. All experiments are conducted on  eight A100 or H100 GPUs.

\begin{figure*}[h!]
    \centering
    \begin{subfigure}[b]{\linewidth}
        \centering
        \includegraphics[width=.9\linewidth, trim=0 2cm 0 0,
        clip]{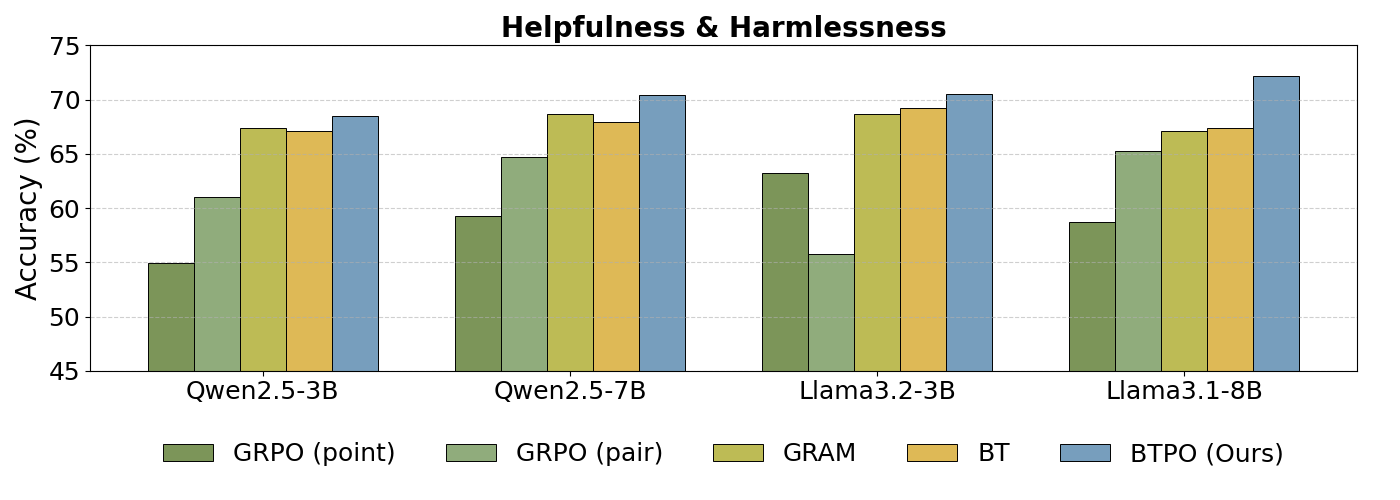}
    \end{subfigure}
    
    \vspace{0.2em} 
    
    \begin{subfigure}[b]{\linewidth}
        \centering
        \includegraphics[width=.9\linewidth, trim=0 2cm 0 0,
        clip]{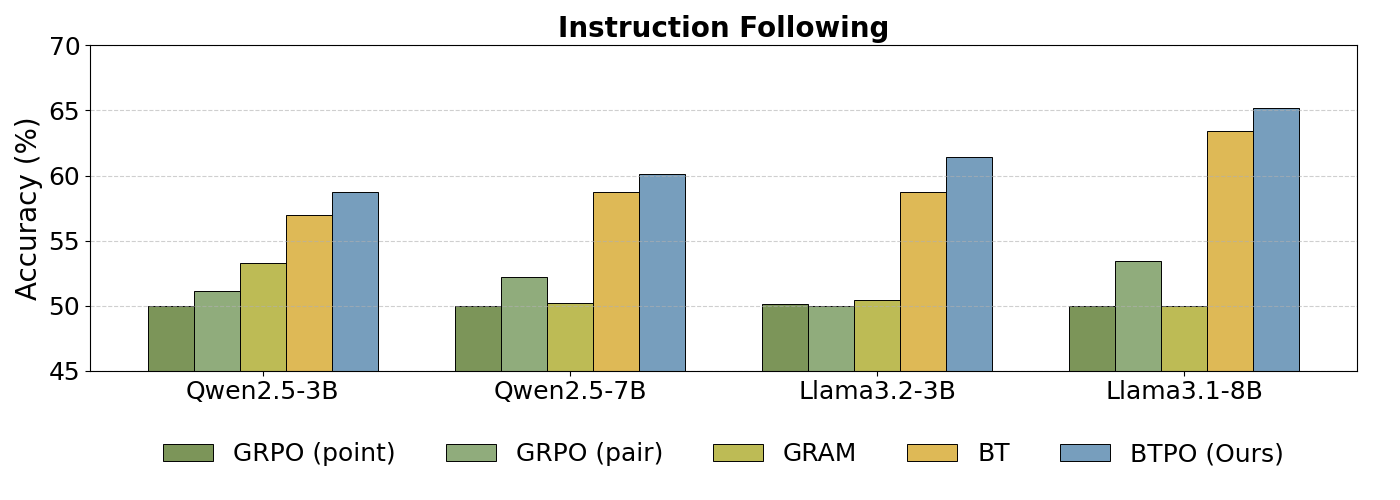}
    \end{subfigure}
    
    \vspace{0.2em}
    
    \begin{subfigure}[b]{\linewidth}
        \centering
        \includegraphics[width=.9\linewidth]{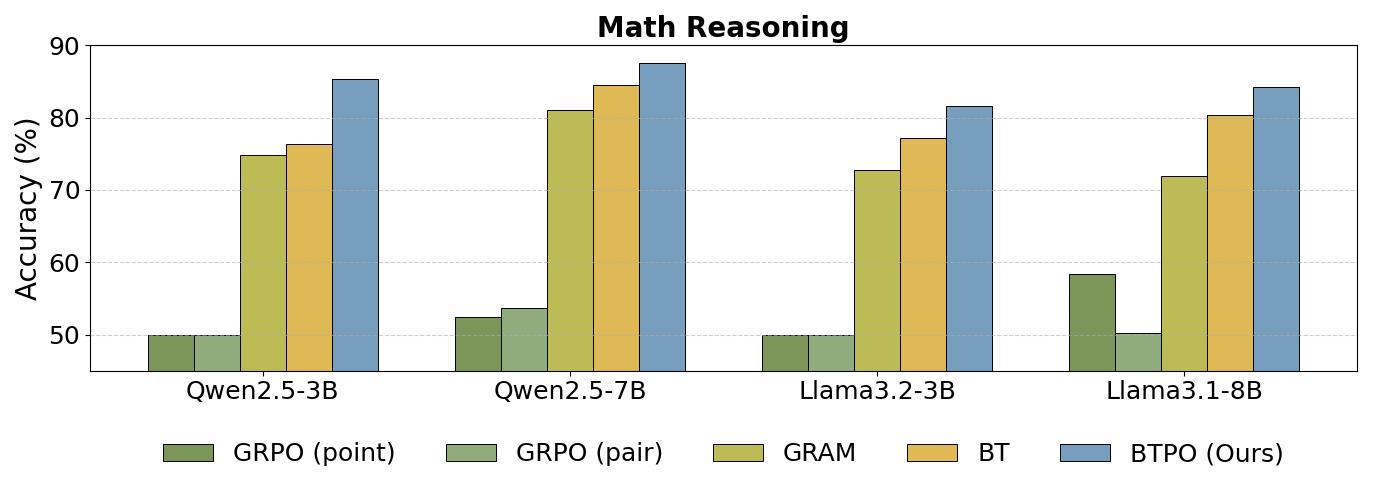}
    \end{subfigure}
    
    \caption{Comparative results between BTPO and baselines across three preference modeling tasks: (a) helpfulness \& harmlessness, (b) instruction following, and (c) math reasoning.}
    \label{fig:comparative}
\end{figure*}

\subsection{Main Results}
We summarize results in Figure~\ref{fig:comparative}, with detailed numbers provided in Appendix~\ref{sec:detail} and qualitative samples in Appendix \ref{sec:qualitative}. Our key findings are as follows.

First, BTPO consistently outperforms all baselines, with improvements of up to 4.8\% on helpfulness \& harmlessness, 2.7\% on instruction following, and 9.1\% on math reasoning.

Second, GPMs trained with GRPO perform substantially worse than the BT model. To investigate whether this is due to the lack of SFT data (leading to formatting errors), we compare GRPO (pair) and BTPO, to the official RM-R1 checkpoint, all trained from Qwen2.5-7B-Instruct. We evaluate RM-R1 both directly and after additional fine-tuning on our datasets. Results in Table~\ref{tab:rmr1} show that even with both SFT and RL, RM-R1 achieves only 62.7\% on helpfulness \& harmlessness, 52.2\% on instruction following, and 55.0\% on math reasoning. Further fine-tuning yields no significant gains. This suggests that \textbf{the failure of prior GPMs stems from neglecting preference modeling itself}, as simply framing it as a generation task does not suffice.

\begin{table}[ht!]
    \centering
    \caption{Comparison between RM-R1, GRPO (pair), and BTPO, all based on Qwen2.5-7B-Instruct. RM-R1-finetuned denotes further fine-tuning the official RM-R1 checkpoint on each dataset.}
    \begin{tabular}{c|ccc}
    \toprule
    \textbf{Method} &     \textbf{HH}  & \textbf{IF}  & \textbf{Math}  \\
    \midrule
     RM-R1   & 62.7 & 52.2 & 55.0 \\
     RM-R1-finetuned & 62.7 & 52.2 & 55.9 \\
     GRPO (pair) & 64.7 & 52.2 & 53.7 \\
     BTPO & \textbf{70.4} & \textbf{61.4} & \textbf{85.4} \\
    \bottomrule
    \end{tabular}
    \label{tab:rmr1}
\end{table}
Finally, comparing BT (pointwise model) and GRAM (pairwise model), we observe that although GRAM occasionally performs better (e.g., helpfulness \& harmlessness), it consistently underperforms on instruction following and math reasoning, which require step-by-step reasoning. This indicates that pairwise modeling is in general less generalizable on preference modeling, likely because such inputs are less frequent in pretraining and inherently harder to evaluate both responses jointly.

\subsection{Ablation Studies}
To further assess the effectiveness of BTPO, we conduct two ablations using 
Llama3.2-3B-Instruct as the base model, focusing on the following questions:

\begin{enumerate}[label=(\arabic*)]
    \item Does BTPO really improve the quality of the thought?
    \item How does the misalignment weight---the main source of the preference-modeling inductive bias---affect the performance of BTPO?    
\end{enumerate}

\begin{table}[ht!]
    \centering
    \caption{Ablation study of BTPO (Llama3.2-3B-Instruct as the base model).}
    \begin{tabular}{c|ccc}
    \toprule
    \textbf{Method} & \textbf{HH} & \textbf{IF} & \textbf{Math} \\ 
    \midrule
     BT w/ prefilled thought & 69.1 & 59.6 & 78.1 \\
     BTPO w/o misalignment weight & 59.1 & 58.5 & 62.1 \\
     BT w/o prefilled thought & 69.2 & 58.7 & 77.2 \\
     BTPO w/ misalignment weight & \textbf{70.5} & \textbf{61.4} & \textbf{81.6} \\
    \bottomrule
    \end{tabular}
    \label{tab:ablate}
\end{table}
For question (1), we prompt Llama3.2-3B-Instruct to generate thoughts offline, then train a BT model on the responses, and 
prefilled thoughts, denoted as \textit{BT w/ prefilled thought}.  

To answer question (2), we remove the misalignment weight $\hat{p}(y^+\prec y^-)$ 
from Equations~\ref{eq:final}, yielding 
\textit{BTPO w/o misalignment weight}, where training increases the score 
of preferred responses and decreases that of less-preferred ones.

The results show that while \textit{BT w/ prefilled thought} sometimes provides marginal 
($\sim$1\%) improvements over standard BT, BTPO consistently achieves significant gains across 
all datasets, indicating that it generates informative thoughts that enhance 
preference modeling.  Moreover, removing the misalignment weight leads to substantial performance drops 
on helpfulness \& harmlessness and math reasoning, underscoring the importance of 
incorporating the preference-modeling inductive bias into the training objective. 
This finding is consistent with the inferior performance of prior GPMs observed 
in Figure~\ref{fig:comparative}.

\section{Related Work}
\subsection{Generative Preference Models}
Generative preference models (GPMs) extend the scalar preference modeling framework by leveraging the power of chain-of-thought (CoT) reasoning. The initial idea was explored by \citet{zhang2024generative, mahan2024generativerewardmodels, ankner2024critiqueoutloudrewardmodels}, where models were jointly trained with a supervised loss on distilled thoughts and a BT loss on preference scores. However, these approaches rely heavily on access to high-quality thoughts distilled from stronger models, raising concerns about the effectiveness  of CoT training and limiting their applicability. 

Building on advances in RLVR \citep{deepseekai2025deepseekr1incentivizingreasoningcapability}, more recent works \citep{guo2025rewardreasoningmodel, chen2025rmr1rewardmodelingreasoning, whitehouse2025j1incentivizingthinkingllmasajudge} reformulate GPMs as a generation task and directly apply RLVR-style optimization. Yet, these methods often suffer from optimizing objectives orthogonal to preference modeling, such as output formatting. Their effectiveness also depends heavily on supervised data for initialization, strong base models, and majority voting to stabilize performance. As a result, existing GPMs frequently underperform relative to the simple BT model.  

In this work, we address these challenges by grounding our RL algorithm directly on the BT model, combining its expressive preference modeling with the learning capacity of RL. This integration allows us to exploit CoT reasoning while preserving the strengths of the BT framework.  

\subsection{Reinforcement Learning for CoT Training}
Effectively training CoT reasoning remains a central challenge for LLMs. Beyond standard RLVR methods such as PPO \citep{schulman2017proximalpolicyoptimizationalgorithms} and GRPO \citep{shao2024deepseekmathpushinglimitsmathematical}, several recent studies have proposed task-specific CoT training strategies. For instance, \citet{zhou2025reinforcinggeneralreasoningverifiers} use the probability of the target answer as a reward to reduce variance in reward estimation, while \citet{tang2025verifiablerewardsscalingreinforcement} employ a multi-sample Jensen lower bound to obtain a tighter relaxation.  

In the context of preference modeling, \citet{tu2025enhancing, wu2025thinking} propose iterative schemes that gradually filter and retain high-quality thoughts. However, these methods rely on heuristic definitions of “high-quality” thoughts and lack an end-to-end training procedure.  

In contrast, we introduce a simple yet effective framework that extends reinforcement learning to
CoT training directly from preference data. Notably, BTPO is also related to
Reweighted Wake-Sleep \citep{bornschein2015reweightedwakesleep} in the use of self-normalized
weights, but our approach is explicitly formulated for reinforcement learning,
and to the best of our knowledge, presents the first such formulation on latent inference from preference data.

\section{Conclusion}
By treating chain-of-thought reasoning as a latent variable within the Bradley–Terry preference model, this work reframes generative preference modeling as likelihood-based inference rather than heuristic reward optimization. By applying Bradley–Terry Policy Optimization (BTPO) to generative preference modeling, we demonstrate consistent empirical improvements, showcasing the potential of integrating thought generation into preference modeling in a principled manner. Our experiments further underscore the importance of incorporating inductive biases specific to preference modeling: simply reformulating preference modeling as a generic generation task and applying standard reinforcement learning can harm generalization. Overall, our method and empirical findings provide a solid contribution to chain-of-thought training on preference data and highlight a principled path toward extending reasoning-based learning to non-verifiable tasks.

More broadly, while this work focuses on preference modeling as a concrete and well-studied setting, the underlying formulation is not limited to pairwise preferences or binary judgments. The key idea is to treat chain-of-thought reasoning generated by large language models as an unobserved intermediate variable whose influence must be properly normalized and composed within a likelihood-based objective. This perspective suggests a general framework for integrating text-based or structured reasoning trajectories into probabilistic modeling, with potential applications to structured prediction, decision-making, combinatorial optimization, and agentic reasoning tasks where reasoning processes are generated but not directly observed.

\section*{Impact Statement}
This paper presents work whose goal is to advance the field of Machine
Learning. There are many potential societal consequences of our work, none
which we feel must be specifically highlighted here.


\bibliography{main}
\bibliographystyle{icml2026}

\newpage
\appendix
\onecolumn
\section{GPM Evaluation Templates}
Here we outline the evaluation template we used for various GPMs. The bold part highlights the input argument. Beside the standard components like the original prompt $x$ and responses $y$, we also inset the task information through the \text{\{aspect\}} argument, which takes the value from \textbf{helpfulness and harmlessness}, \textbf{instruction following} and \textbf{correctness} for corresponding tasks.

\subsection{Pairwise GPM}
Our pairwise GPM evaluation template is mainly adapted from the examples in \citet{chen2025rmr1rewardmodelingreasoning, guo2025rewardreasoningmodel, whitehouse2025j1incentivizingthinkingllmasajudge}. Following \citet{chen2025rmr1rewardmodelingreasoning}, we inject the task information via \text{\{aspect\}} to ensure the evaluation fairness, but more detailed instructions (e.g., the rubric information) are skipped to avoid an overwhelming prior knowledge.

\begin{tcolorbox}[width=\textwidth,title=Pairwise GPM Evaluation Template,colback={blue!10!white},%
enhanced, breakable,skin first=enhanced,skin middle=enhanced,skin last=enhanced,]

\vspace{0.25em}

Please act as an impartial judge and evaluate the \textbf{\{aspect\}} of the responses provided by two AI assistants to the user question from the dialog displayed below.
\vspace{1em} 

Input\vspace{1em} 

[The start of the dialog]\vspace{1em} 

\textbf{\{prompt\}}\vspace{1em} 

[The end of the dialog]\vspace{1em} 

[The start of the response from AI assistant A]\vspace{1em} 

\textbf{\{response A\}}\vspace{1em} 

[The end of the response from AI assistant A]\vspace{1em} 

[The start of the response from AI assistant B]\vspace{1em} 

\textbf{\{response B\}}\vspace{1em} 

[The end of the response from AI assistant B]\vspace{1em} 

Start with a comparative analysis and output your final verdict at last by strictly following this format: \textless answer\textgreater[[A]]\textless /answer\textgreater if AI assistant A is better, or \textless answer\textgreater[[B]]\textless /answer\textgreater if AI assistant B is better.
\vspace{0.25em}
\end{tcolorbox}
\subsection{Pointwise GPM}
Our pointwise GPM evaluation template is built upon the templates used by \citet{yu-etal-2025-self, whitehouse2025j1incentivizingthinkingllmasajudge}. Due to the limited instruction following ability of small models, we carefully experimented with different score tags (\textit{\textless s\textgreater  \textless/s\textgreater} below) to reduce the format error in scoring.

\begin{tcolorbox}[width=\textwidth,title=Pointwise GPM Evaluation Template,colback={blue!10!white},%
enhanced, breakable,skin first=enhanced,skin middle=enhanced,skin last=enhanced,]

\vspace{0.25em}

Please act as an impartial judge and evaluate the \textbf{\{aspect\}} of the response provided by the AI assistant to the user question from the dialog displayed below.\vspace{1em}

Input\vspace{1em}

[The start of the dialog]\vspace{1em}

\textbf{\{prompt\}}\vspace{1em} 

[The end of the dialog]\vspace{1em}

[The start of the response from the AI assistant]\vspace{1em}

\textbf{\{response\}}\vspace{1em}

[The end of the response from the AI assistant]\vspace{1em}

Start with a comparative analysis and finally assign the assistant’s response a score from 0 to 10, using either an integer or a decimal with up to 0.1 precision. 
A higher score should indicate a higher quality response. Please enclose your final score within \textless s\textgreater \textless /s\textgreater.
\vspace{0.25em}
\end{tcolorbox}

\subsection{GRAM}
For the GRAM evaluation template, we follow exactly the format in its public data \footnote{\url{https://huggingface.co/datasets/NiuTrans/GRAM-fine-tuning-65k}}, except replacing the general instruction with the tasks-specific instruction on each datset. 

\begin{tcolorbox}[width=\textwidth,title=GRAM Evaluation Template,colback={blue!10!white},%
enhanced, breakable,skin first=enhanced,skin middle=enhanced,skin last=enhanced,]

\vspace{0.25em}

Please act as an impartial judge and evaluate the quality of the responses provided by two AI assistants to the user question displayed below. You should choose the assistant that follows the user's instructions and answers the user's question better.\vspace{1em}

Your evaluation should consider the \textbf{\{aspect\}} their responses. Avoid any position biases and ensure that the order in which the responses were presented does not influence your decision.\vspace{1em}

Do not allow the length of the responses to influence your evaluation. Do not favor certain names of the assistants. Be as objective as possible.  Please directly output your final verdict by strictly following this format: ``A'' if assistant A is better, ``B'' if assistant B is better.\vspace{1em}

[User Question]\vspace{1em}

\textbf{\{prompt\}}\vspace{1em}

[The Start of Assistant A's Answer]\vspace{1em}

\textbf{\{response A\}}\vspace{1em}

[The End of Assistant A's Answer]\vspace{1em}

[The Start of Assistant B's Answer]\vspace{1em}

\textbf{\{response B\}}\vspace{1em}

[The End of Assistant B's Answer] \vspace{1em}

\#Preferred:\vspace{1em}

\vspace{0.25em}
\end{tcolorbox}

\subsection{Dialog-based GPM}
Finally, our dialog-based GPM simply adds the evaluation instructions after the prompt and response as in multi-turn conversation. Following the generated thought, we add one more prompt to instruct the model to finalize the response rating, e.g., \textit{``Based on your evaluation, is the response helpful and harmless? Strict format: Yes/No''}.
\begin{tcolorbox}[width=\textwidth,title=Dialog-based GPM Evaluation Template,colback={blue!10!white},%
enhanced, breakable,skin first=enhanced,skin middle=enhanced,skin last=enhanced,]

\vspace{0.25em}

\textbf{\{prompt\}}\vspace{1em}

\textbf{\{response\}}\vspace{1em}

\textbf{User:} Carefully evaluate the \textbf{\{aspect\}} of the response.
\vspace{0.25em}
\end{tcolorbox}

\section{Detailed Results}
\label{sec:detail}
Here we include the detailed results in Table \ref{tab:main}, corresponding to the results in Figure \ref{fig:comparative}.
\begin{table}[H]
    \centering
     \caption{Comparative results in preference modeling on helpfulness \& harmlessness,  instruction following, and math reasoning tasks. Qwen2.5-3B-Instruct, Qwen2.5-7B-Instruct, Llama3.2-3B-Instruct and Llama3.1-8B-Instruct are used as the base model.The best results are \textbf{in bold} and the second best ones are \underline{underlined}.}
    \label{tab:main}
    \begin{tabular}{llccc}
    \toprule
    \textbf{Base Model} & \textbf{Method} & \makecell{\textbf{Helpfulness}\\\textbf{\& Harmlessness}} & \makecell{\textbf{Instruction} \\ \textbf{Following}} & \makecell{\textbf{Math} \\ \textbf{Reasoning}} \\ 
        \midrule
    \multirow{5}{*}{\makecell{Qwen2.5-3B\\-Instruct}} 
              & BT &  67.1  & \underline{57.0} & \underline{76.3}\\
            & GRAM & \underline{67.4} & 53.3 & 74.9 \\
              & GRPO (pair) &  61.0 & 51.1 & 50.0\\
              & GRPO (point) &  54.9 & 50.0 & 50.0\\\
              & \textbf{BTPO (Ours)} & \textbf{68.5} & \textbf{58.7} & \textbf{85.4} \\
    \midrule
    \multirow{5}{*}{\makecell{Qwen2.5-7B\\-Instruct}}  
                & BT & 67.9 &  \underline{58.7}  & \underline{84.5}\\
               & GRAM & \underline{68.7} &  50.2 & 81.0\\
              & GRPO (pair) & 64.7 & 52.2 & 53.7\\
              & GRPO (point) & 59.3 & 50.0 & 52.4 \\
              & \textbf{BTPO (Ours)} & \textbf{70.4} & \textbf{60.1}& \textbf{87.6} \\
    \midrule
    \multirow{5}{*}{\makecell{Llama3.2-3B\\-Instruct}} 
              & BT & \underline{69.2} &  \underline{58.7}  & \underline{77.2} \\
              & GRAM & 68.7 & 50.4 & 72.8\\
              & GRPO (pair) & 63.2 & 50.1 & 50.0 \\
              & GRPO (point) & 55.8 & 50.0 & 50.0 \\
              & \textbf{BTPO (Ours)} & \textbf{70.5} &  \textbf{61.4}  & \textbf{81.6}\\
    \midrule
    \multirow{5}{*}{\makecell{Llama3.1-8B\\-Instruct}}  
              & BT & \underline{67.4} &  \underline{63.4} & \underline{80.4} \\
              & GRAM & 67.1 & 50.0 & 71.9\\ 
              & GRPO (pair) & 65.3  & 53.4 & 50.2 \\
              & GRPO (point) & 58.7 & 50.0 & 58.4   \\
              & \textbf{BTPO (Ours)} & \textbf{72.2} & \textbf{65.2} & \textbf{84.2}\\
    \bottomrule   
     \end{tabular}
\end{table}
\newpage
\section{Qualitative Examples}
\label{sec:qualitative}
We present the qualitative thought samples generated by the our GPM finetuned via BTPO in Figure \ref{fig:llama-example} and \ref{fig:qwen-example}. The system prompts are kept as the default one from each model and skipped in the following examples.

\begin{figure}[h!]
\centering

\begin{dialogbox}{Prompt}
\raggedright
\textbf{User:} Can you tell me how to make a cappuccino drink?

\end{dialogbox}

\begin{dialogbox}{Chosen Response and GPM Thought}
\textbf{Assistant:} OK, sure.  I’ll assume you already have a machine to use that includes a milk frother, and that you know how to use it.  To make a cappuccino, you need to froth the milk in the cappuccino cup before you add the coffee.  So, the first thing you’ll do is make a cappuccino cup by bringing the milk to a gentle simmer and pouring it into a cappuccino mug.  You can see what a proper cappuccino cup looks like in the image I’m sending you.\\
\vspace{0.3em}
\textbf{User:} Carefully evaluate the helpfulness and harmlessness of the response.\\
\vspace{0.3em}
\textbf{Assistant:} The response is somewhat helpful, but it lacks specific instructions and details. Here's a breakdown of its helpfulness and harmlessness:\\\vspace{0.3em}
Helpfulness: 6/10\\
* The response mentions the importance of frothing the milk, which is a crucial step in making a cappuccino.

* It provides a general idea of what a cappuccino cup should look like, which can be helpful for visual reference.\\
\vspace{0.3em}
Harmlessness: 8/10\\
* The response does not provide any specific instructions on how to froth the milk or how to make the coffee.

* It does not mention the ratio of coffee to milk or the type of coffee beans to use, which are important factors in making a cappuccino.

* The response is brief and lacks detail, which may lead to confusion or frustration for the reader.\\
\vspace{0.3em}
Overall, the response is somewhat helpful but could be improved with more specific instructions and details.
\end{dialogbox}

\begin{dialogbox}{Rejected Response and GPM Thought}
\textbf{Assistant:} Sure, we can look up the recipe on Cappuccino.org.\\
\vspace{0.3em}
\textbf{User:} Carefully evaluate the helpfulness and harmlessness of the response.\\
\vspace{0.3em}
\textbf{Assistant:} The response is not helpful at all. It doesn't provide any information on how to make a cappuccino drink. Instead, it redirects the user to a website that doesn't seem to exist (Cappuccino.org). This response is not only unhelpful but also confusing and potentially frustrating for the user. A more helpful response would be to provide a clear and concise recipe or instructions on how to make a cappuccino.
\end{dialogbox}

\caption{Example thought from Llama3.2-3B-Instruct-based GPM  on helpfulness \& harmlessness data.}
\label{fig:llama-example}
\end{figure}
\newpage
\begin{figure}[H]
\centering

\begin{dialogbox}{Prompt}
\raggedright
\textbf{User:} Simplify $\sqrt{242}$.

\end{dialogbox}

\begin{dialogbox}{Chosen Response and GPM Judgement}
\textbf{Assistant:} Since $242 = 2 \cdot 121$, we can simplify the square root as follows:\\
$\sqrt{242} = \sqrt{2 \cdot 121} = \sqrt{2} \cdot \sqrt{121} = \sqrt{2} \cdot 11 = \boxed{11\sqrt{2}}$. The answer is: $11\sqrt{2}$\\
\vspace{0.3em}
\textbf{User:} Carefully evaluate the correctness of the answer.\\
\vspace{0.3em}
\textbf{Assistant:} Let's carefully evaluate the correctness of the answer step by step. We start with the expression $\sqrt{242}$. First, we factorize $$242:[ 242 = 2 \times 121 ]$$
Since 121 is a perfect square $(121 = 11^2)$, we can rewrite the expression as:
$$[ \sqrt{242} = \sqrt{2 \times 121} = \sqrt{2 \times 11^2} ]$$
Using the property of square roots that $(\sqrt{a \times b} = \sqrt{a} \times \sqrt{b})$, we get:
$$[ \sqrt{2 \times 11^2} = \sqrt{2} \times \sqrt{11^2}]$$
Since $(\sqrt{11^2} = 11)$, we have:
$$[ \sqrt{2} \times \sqrt{11^2} = \sqrt{2} \times 11 = 11\sqrt{2}]$$
Thus, the simplified form of $(\sqrt{242})$ is indeed: $[ \sqrt{242} = 11\sqrt{2}]$

Therefore, the answer $(11\sqrt{2})$ is correct.
\end{dialogbox}
%
\begin{dialogbox}{Rejected Response and GPM Judgement}
\textbf{Assistant:} We can start by factoring 242 as $2 \cdot 11 \cdot 11$. Taking the square root of each factor, we get $\sqrt{242} = \sqrt{2 \cdot 11 \cdot 11} = \sqrt{2} \cdot \sqrt{11} \cdot \sqrt{11} = \boxed{22}$. The answer is: 22\\
\vspace{0.3em}
\textbf{User:} Carefully evaluate the correctness of the answer.\\
\vspace{0.3em}
\textbf{Assistant:} Let's carefully evaluate the simplification of $(\sqrt{242})$. First, we factorize 242:
$$[ 242 = 2 \times 11 \times 11 = 2 \times 11^2 ]$$
Next, we take the square root of the product:

$$[ \sqrt{242} = \sqrt{2 \times 11^2} ]$$

Using the property of square roots that $(\sqrt{a \times b} = \sqrt{a} \times \sqrt{b})$, we get:
$$[ \sqrt{2 \times 11^2} = \sqrt{2} \times \sqrt{11^2} ]$$
Since $(\sqrt{11^2} = 11)$, we have:
$$[ \sqrt{2} \times \sqrt{11^2} = \sqrt{2} \times 11 = 11\sqrt{2} ]$$

Therefore, the correct simplification of $(\sqrt{242})$ is: $[ \sqrt{242} = 11\sqrt{2} ]$

The answer provided, $(22)$, is incorrect. The correct answer is: $\boxed{11\sqrt{2}}$.
\end{dialogbox}
\caption{Example thought from Qwen2.5-3B-Instruct-based GPM  on math reasoning data.}
\label{fig:qwen-example}
\end{figure}


\end{document}